\documentclass[10pt,twocolumn,letterpaper]{article}

\usepackage[pagenumbers]{cvpr} 

\usepackage[dvipsnames]{xcolor}

\usepackage{kotex}
\usepackage{multirow}
\usepackage{algorithm,algorithmic}

\usepackage{xcolor, color, colortbl}

\definecolor{dark_cyan}{RGB}{0,139,139}
\definecolor{OliveGreen}{rgb}{0.5, 0.5, 0.0}
\definecolor{Fuchsia}{rgb}{1.0, 0.0, 1.0}

\newcommand{\nj}[1]{\textcolor{black}{#1}}
\newcommand{\hj}[1]{\textcolor{black}{#1}}

\usepackage[accsupp]{axessibility}

\usepackage{chngcntr} 

\definecolor{cvprblue}{rgb}{0.21,0.49,0.74}
\usepackage[pagebackref,breaklinks,colorlinks,citecolor=cvprblue]{hyperref}

\def\paperID{8} 
\def\confName{CVPR}
\def\confYear{2024}

\title{Coreset Selection for Object Detection}

\author{Hojun Lee$^1$ \quad Suyoung Kim$^1$ \quad Junhoo Lee$^1$ \quad Jaeyoung Yoo$^2$ \quad Nojun Kwak$^{1}$\thanks{Corresponding author} \\
$^1$Seoul National University \qquad \ \ \ \ \ $^2$NAVER WEBTOON AI\\
{\tt\small \{hojun815,ksyo96,mrjunoo,nojunk\}@snu.ac.kr, yoojy31@webtoonscorp.com}
}

\begin{document}
\maketitle
\begin{abstract}
Coreset selection is a method for selecting a small, representative subset of an entire dataset. It has been primarily researched in image classification, assuming there is only one object per image. 
However, coreset selection for object detection is more challenging as an image can contain multiple objects. As a result, much research has yet to be done on this topic. 
Therefore, we introduce a new approach, \textit{\textbf{C}oreset \textbf{S}election for \textbf{O}bject \textbf{D}etection} (CSOD).
CSOD generates imagewise and classwise representative feature vectors for multiple objects of the same class within each image. Subsequently, we adopt submodular optimization for considering both representativeness and diversity and utilize the representative vectors in the submodular optimization process to select a subset. 
When we evaluated CSOD on the Pascal VOC dataset, CSOD outperformed random selection by +6.4\%p in AP$_{50}$ when selecting 200 images.
\end{abstract}    
\section{Introduction}
\label{sec:intro}

\begin{figure}[t]
    \centering
    \includegraphics[width=0.99\linewidth]{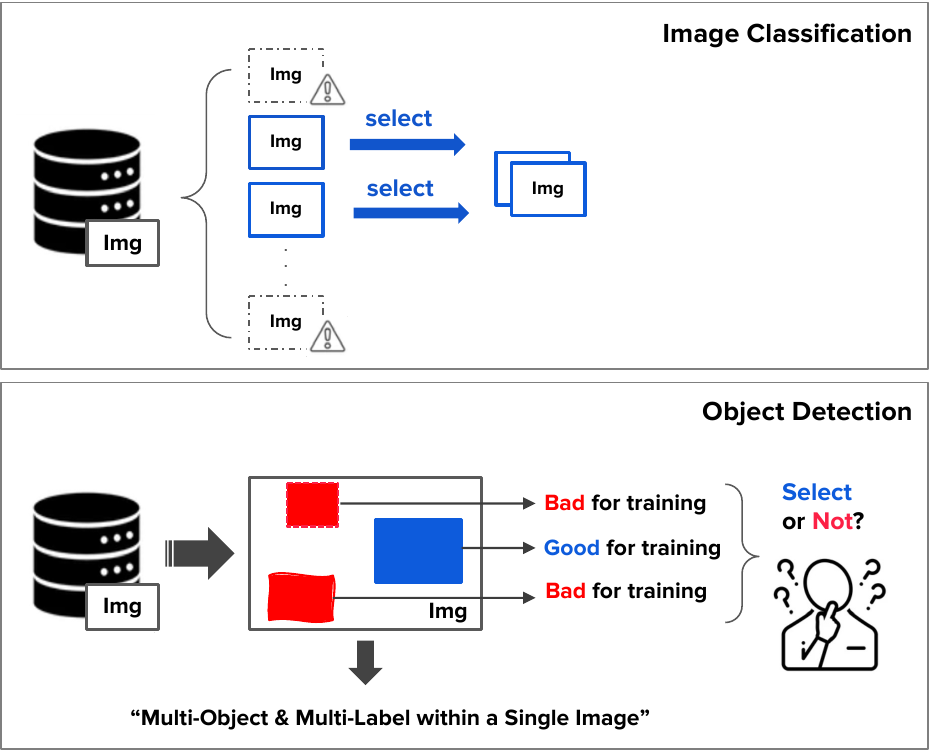}
    \caption{The difference in coreset selection between image classification and object detection.}
    \label{fig:overview}
\end{figure}

In today's data-driven era, managing the sheer volume and variety of data presents a crucial challenge, particularly in areas like computer vision and deep learning, which deal with a tremendous amount of data~\cite{radford2021learning,codetr2022,cheng2021mask2former}.
With the advent of technologies such as autonomous vehicles and smart surveillance systems, accurate and efficient recognition of such image data has become paramount. One key strategy in managing these massive datasets involves `coreset selection,' a method aimed at identifying a smaller, representative subset of the original dataset. This subset is then used to streamline complex computations and enhance processing efficiency.

However, as illustrated in Figure \ref{fig:overview}, traditional coreset selection methods are unrealistic because they were developed under the naïve assumption of a single object per image, a condition that real-world images do not often meet~\cite{braverman2019coresets,coleman2019selection,guo2022deepcore}. Real-world images typically contain the natural variability, such as multiple objects of various categories, sizes, and locations.

This implies that we should develop methods that consider the natural variability. As Figure \ref{fig:overview} illustrates, if we evaluate the suitability of an image on an object-by-object basis, considering one object as suitable for training does not necessarily imply that the others are also suitable. In other words, the
decision of a core image should be based on all objects in an image.
However, traditional methods, designed to solve only the single-image-single-object pairing, fail to consider this, and therefore struggle in realistic conditions.

\hj{In this paper, we address a significant limitation in the field of coreset selection which has traditionally operated under the assumption of a single object per image. We introduce a realistic approach tailored to the more complex, yet common, scenario where images inherently contain multiple objects. This shift from single-object to multi-object consideration is a central advancement of our work.}
CSOD not only recognizes the presence of numerous objects within each image but also tackles the compounded uncertainties by considering objects' spatial information such as size and location. To validate our method, we implemented and conducted experiments in object detection, a representative task of situations where multiple objects may reside within a single image.

Our method, CSOD, is built upon a unique concept: the `imagewise-classwise vector.' To select the most representative images among compounded uncertainties, we need a way to summarize the information of each image effectively. The imagewise-classwise vector serves this purpose by averaging the features of objects of the same class within an image. This comprehensive representation allows for informed decision-making when addressing the complexities of multi-object images.

We employ a greedy approach to select individual data points sequentially by class order, thereby constructing the coreset step by step. Although this method considers only one class at each selection step, it guarantees that the most pertinent selections for each class are made, enhancing the representativeness and diversity of the coreset. Furthermore, to ensure that the selected subset informatively represents the entire dataset, we introduce a mathematical tool known as a `submodular function,' as delineated by \citet{krause2014submodular}. The function aids in selecting the most informative subset based on the imagewise-classwise average features for each category. 

Our empirical evaluations, particularly in scenarios involving the detection of multiple objects, demonstrate the effectiveness of CSOD. For instance, when selecting 200 images from the Pascal VOC dataset~\cite{everingham2007pascal}, our method achieved an impressive improvement of +6.4\% point in AP$_{50}$ compared to random selection. Moreover, we also evaluated it on the BDD100k~\cite{yu2020bdd100k} and MS COCO2017~\cite{lin2014microsoft} datasets and confirmed that our method outperforms random selection. These significant achievements emphasize the efficacy and innovativeness of CSOD in addressing the challenges of coreset selection in multi-object image data.

\hj{In summary, CSOD represents a pivotal extension of existing coreset selection frameworks to encompass scenarios with multiple objects per image. This approach addresses a gap that traditional methods, which assumed only one object per image, did not cover.}
While our focus is on multi-objects of the same class, 
we acknowledge the potential for future expansions of our method to accommodate images featuring various categories of objects. 
With our unique problem recognition and solution, 
we aim to shift the paradigm of coreset selection towards more realistic scenarios, specifically image datasets containing multiple objects.
This research transcends mere technical advancement, marking a pivotal shift in processing complex real-world datasets and opening new horizons for coreset selection, thereby addressing a significant challenge of the big data era.
\section{Background and Prior works}
\label{sec:background}

\subsection{Coreset selection}
\label{subsec:background_coresetselection}
\citet{welling2009herding} introduced the concept of herding for iterative data point selection near class centers.
\citet{wei2015submodularity} applied the submodular function to the Naïve Bayes and Nearest Neighbor classifier.
We also adopt this function, so we provide further explanation in Section~\ref{subsec:background_submodular}. \citet{braverman2019coresets,huang2019coresets} modified statistical clustering algorithms like k-median and k-means to identify data points that effectively represent the dataset. \citet{coleman2019selection} utilized uncertainties measured by entropy or confidence. \citet{huang2023coresets} theoretically explained the upper and lower bounds on the coreset size for k-median clustering in low-dimensional spaces.
However, most previous researches focused on image classification, and to the best of our knowledge, our work is the first research to design coreset selection specifically for object detection.

\subsection{Dataset Distillation}
\label{subsec:background_datasetdistill}
Coreset Selection and Dataset Distillation are crucial in enhancing model training efficiency, with the former selecting informative data points and the latter synthesizing data to distill the dataset's information. Despite their different approaches—selection versus synthesis—both methods aim to encapsulate data. Current Dataset Distillation research~\cite{wang2018dataset,nguyen2021dataset,cazenavette2022dataset,zhou2022dataset,dong2022privacy}, primarily focused on image classification, presents unexplored potential in object detection. Advancements in Coreset Selection for object detection may 
 have a significant influence on Dataset Distillation strategies for object detection.

\subsection{Submodular function}
\label{subsec:background_submodular}
A set function $\displaystyle f:2^{\mathcal{V}} \rightarrow \mathbb{R}$ is considered submodular if, for any subsets $\mathcal{A}$ and $\mathcal{B}$ of $\mathcal{V}$ where $\mathcal{A} \subseteq \mathcal{B}$ and $\displaystyle x$ is an element not in $\mathcal{B}$, the following inequality holds:
\begin{equation}
f(\mathcal{A} \cup \{x\}) - f(\mathcal{A}) \geq f(\mathcal{B} \cup \{x\}) - f(\mathcal{B})
\end{equation}
Here, $\Delta(x|\mathcal{A}) := f(\mathcal{A} \cup \{x\})-f(\mathcal{A})$ represents the benefit of adding $\displaystyle x$ to the set $\mathcal{A}$. In simple terms, this inequality means that adding $x$ to $\mathcal{B}$ provides less additional benefit than adding $x$ to $\mathcal{A}$. This is because $\mathcal{B}$ already contains some of the information that $x$ can offer to $\mathcal{A}$. Therefore, we can use submodularity to find a subset that maximizes the benefit of adding each element.

However, in general, selecting a finite subset $\mathcal{S}$ with the maximum benefit is a computationally challenging problem (NP-hard) \citep{krause2014submodular}. 
To address this, we employ a greedy algorithm that starts with an empty set and adds one element at a time. Specifically, $\mathcal{S}_{i}$ is updated as $\mathcal{S}_{i-1} \cup \operatorname*{argmax}_x \Delta (x | \mathcal{S}_{i-1})$. For more information, please refer to \citet{krause2014submodular}.

\subsection{Faster R-CNN}
\label{subsec:background_frcnn}
\begin{figure}
    \centering
    \includegraphics[width=0.965\linewidth]{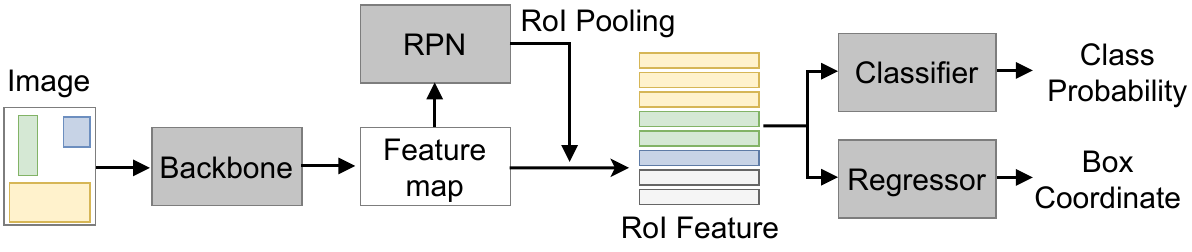}
    \vspace{-1.5mm}
    \caption{The forward process during the training phase of Faster R-CNN. The RoI features include both foreground and background regions at the forward process.}
    \label{fig:frcnn}\vspace{-1.1mm}
\end{figure}

Various 
object detectors exist, including Faster R-CNN \citep{ren2015fasterRCNN}, SSD \citep{fu2017dssd}, YOLO \citep{redmon2018yolov3}, and DETR \citep{carion2020detr}.
We chose Faster R-CNN as our base model. This choice was motivated by its widespread adoption not only in supervised detection but also in various research areas such as few-shot detection \citep{wang2020frustratingly}, continual learning \cite{wang2021wanderlust}, and semi-supervised object detection \citep{jeong2019consistency}.

Faster R-CNN operates as a two-stage detector. As illustrated in Figure \ref{fig:frcnn}, the first stage employs Region Proposal Network (RPN) to generate class-agnostic object candidate regions in the image, followed by pooling these regions to obtain Region of Interest (RoI) feature vectors. In the second stage, the model utilizes these RoI feature vectors for final class prediction and bounding box regression. Our research uses these RoI feature vectors for coreset selection.

\subsection{Active Learning for Object Detection}
\label{subsec:background_active_for_od}
Active learning is concerned with selecting which unlabeled data to annotate and is thus related to coreset selection. In the context of active learning for object detection, \citet{yuan2021multiple} proposed a method based on uncertainty that utilizes confidence scores on unlabeled data. 
\citet{kothawade2022talisman} aimed to address the low performance issue in rare classes when conducting active learning. The method extracted features of rare classes from labeled data and aimed to maximize the information of rare classes by submodular function and computing the cosine similarity between these labeled features and the features of unlabeled data.

\section{Method}
\label{sec:method}
\subsection{Problem Setup}
\label{sec:problum_setup}
We have an entire training dataset $\mathcal{T}$  = $\{x_i, y_i\}^D_{i=1}$. Here, $x_i \in \mathcal{X} $ is an input image, and  $y_i \in \mathcal{Y}$ is a ground truth. 
Because these data are for object detection, \hj{$y_i=\{c_{i,j}, b_{i,j}\}_{j=1}^{G_i}$}
contains variable numbers of annotations depending on the image. In the \hj{$G_i$} annotations, \hj{$c_{i,j}$} is a class index, and \hj{$b_{i,j}=\{b_{i,j}^{left}, b_{i,j}^{top}, b_{i,j}^{right}, b_{i,j}^{bottom}\}$} denotes the coordinates of the $j$-th bounding box.
Coreset selection aims to choose a \hj{labeled} subset $\mathcal{S} \subset \mathcal{T}$ that best approximates the performance of a model trained on the entire \hj{labeled} dataset, $\mathcal{T}$. 

In our approach, we prioritize the number of images over the number of annotations. This is because annotations typically consist of relatively few strings, and what primarily affects training time and data storage is the number of images rather than the number of annotations.

\begin{algorithm}[t]
\begin{algorithmic}[1]
\footnotesize
\caption{\hj{CSOD Pseudocode}}\label{alg:overall}
\REQUIRE Training Data $\mathcal{T} = \{(x_i, y_i)\}_{i=1}^{D}$ with $C$ classes, where $y_i = \{(c_{i,j}, b_{i,j})\}_{j=1}^{G_i}$ and $G_i$ is the number of ground truth objects in the $i$-th image. Trained backbone $f_{\theta}$. RoI pooler $g$, Global Average Pooling function $h$.  
\ENSURE Selected subset $\mathcal{S}$ with size $N$
\STATE Initialize $\mathcal{S}_{c} = \emptyset$, $\mathcal{P}_{c} = \emptyset$, $\mathcal{Q}_{c} = \emptyset$ for all $c \in \{1, \dots, C\}$
\STATE \textbf{Stage 1}: Preparing Imagewise-Classwise Features
\FOR{$i=1$ to $D$}
    \STATE RoI features $\mathcal{R}_i = \{\mathrm{\boldsymbol{r}}_{i,j}\}_{j=1}^{G_i} = h(g(f_{\theta}(x_i), y_i))$  \hfill\COMMENT{$\triangleright$ Sec. \ref{subsec:method_roi_feat_extract}}
    \FORALL{classes $c$ present in $y_i$}
        \STATE $\mathrm{\boldsymbol{p}}_{i,c} = \frac{1}{|\{j|c_{i,j}=c\}|} \sum_{\{j|c_{i,j}=c\}} \mathrm{\boldsymbol{r}}_{i,j}$ \hfill\COMMENT{$\triangleright$ Sec. \ref{subsec:method_imagewise}}
        \STATE Update $\mathcal{P}_c = \mathcal{P}_c \cup \{\mathrm{\boldsymbol{p}}_{i,c}\}$
    \ENDFOR
\ENDFOR
\STATE \textbf{Stage 2}: Subset Selection
\WHILE{$|\mathcal{S}| < N$}
    \FOR[\hfill $\triangleright$ Sec. \ref{subsec:method_greedy_selection}]{$c=1$ to $C$}
        \STATE Compute scores $s_{i,c} = \text{score}(\mathrm{\boldsymbol{p}}_{i,c}, \mathcal{Q}_{c}), \forall i$ \hfill\COMMENT{$\triangleright$ Eq.\ref{eq:submodular_gain}}
        \STATE Select the image $i^* = \arg\max_i s_{i,c}$
        \STATE Update $\mathcal{S}_{c} = \mathcal{S}_{c} \cup \{(x_{i^*}, y_{i^*})\}$
        \STATE Update $\mathcal{Q}_{c'} = \mathcal{Q}_{c'} \cup \{\mathrm{\boldsymbol{p}}_{i^*,c'}\}, \forall c' \in y_{i^*}$
        \STATE Remove $\mathrm{\boldsymbol{p}}_{i^*,c'}$ from $\mathcal{P}_{c'}$, $\forall c' \in y_{i^*}$
    \ENDFOR
    \STATE Update $\mathcal{S} = \bigcup_{c \in \{1, \dots, C\}} \mathcal{S}_c$
\ENDWHILE
\end{algorithmic}
\vspace{-0.1mm}
\end{algorithm}

\subsection{\hj{Overview}}


CSOD picks out the most useful images by looking at one object category at a time. \nj{Below are the steps of our CSOD}:

\noindent\textbf{Preparing Object Features}: We extract RoI feature vectors from the ground truth of the entire training set (Sec. \ref{subsec:method_roi_feat_extract}). \nj{Then, we average} the RoI features of the same class within one image (Sec. \ref{subsec:method_imagewise}).

\noindent\textbf{Choosing the Best Images}: \nj{We} utilize the averaged RoI feature vectors to greedily select images one by one for each class in a rotating manner (Sec. \ref{subsec:method_greedy_selection}). 
\nj{In doing so, the submodular optimization technique is introduced} to ensure that the selection process considers both representativeness and diversity (Eq.~\ref{eq:submodular_gain}). 
When we pick an image, we do not just use one object in it for training; we use all the \nj{objects it contains}.

Algorithm \ref{alg:overall} provides the pseudocode, while Figure \ref{append_fig:overview} in the supplementary material aids understanding.

\subsection{Ground Truth RoI Feature Extraction}
\label{subsec:method_roi_feat_extract}
With Faster R-CNN, we extract RoI feature vectors from training images by the ground truth (not from \hj{the RPN output).}
If the $i$-th training image contains \hj{$G_i$} ground truth objects, then we have \hj{$G_i$} RoI feature vectors, $\mathcal{R}_i$, as follows:
\begin{equation}
\mathcal{R}_i = \hj{\{\mathrm{\boldsymbol{r}}_{i,j}\}_{j=1}^{G_i}} = h(\displaystyle g(\displaystyle f_{\theta}(x_i), y_i))
\end{equation}
where $x_i$ is an input image, $y_i$ is a ground truth, $f_{\theta}$ is the backbone trained by the entire data, $g$ is the RoI pooler, $h$ is global average pooling and $\mathrm{\boldsymbol{r}}_{i,j}$ is the $j$-th RoI feature vector of the $i$-th image.

\subsection{Imagewise and Classwise Average}
\label{subsec:method_imagewise}
Once we have extracted all the RoI feature vectors for each image, we have a choice to make: For coreset selection, should we average the RoI feature vectors of the same class within a single image to create a single prototype vector representing that class for the image, or should we use these RoI feature vectors directly?

As mentioned in Section \ref{sec:intro}, we chose the averaging approach. If $\mathcal{R}_i = \hj{\{\mathrm{\boldsymbol{r}}_{i,j}\}_{j=1}^{G_i}}$ represents the RoI feature vectors for the $i$-th data with \hj{$G_i$} ground truth objects, then the average RoI feature vector for class $c$ in the $i$-th data, denoted as $\mathrm{\boldsymbol{p}}_{i,c}$, is calculated as follows:
\begin{equation}
\mathrm{\boldsymbol{p}}_{i,c} = \frac{1}{\hj{|\{j|c_{i,j}=c\}|}} \sum_{\{\hj{j|c_{i,j}=c}\}} \mathrm{\boldsymbol{r}}_{i,j}
\end{equation}

\subsection{Greedy Selection}
\label{subsec:method_greedy_selection}
After obtaining averaged RoI feature vectors, our selection process follows a greedy approach, iteratively choosing one data point from each class at a time. To facilitate this, we compute a similarity-based score for each RoI feature vector. This scoring mechanism based on the submodular function assigns higher scores to RoI feature vectors that are similar to others within the same class and lower scores to those similar to RoI feature vectors that have already been selected. This strategy enables us to take into account previously selected data points when making new selections.

\hj{The score function computes the score $s$ for the $i$-th data point within class $c$ as follows:}
\begin{equation}\label{eq:submodular_gain}
    s_{i,c} = \lambda \cdot \sum_j cos(\mathrm{\boldsymbol{p}}_{i,c},\mathrm{\boldsymbol{p}}_{j,c}) - \sum_j  cos(\mathrm{\boldsymbol{p}}_{i,c}, \mathrm{\boldsymbol{q}}_{j,c})
\end{equation}
The term ``$cos$'' represents the cosine similarity, $\mathrm{\boldsymbol{p}}_i$ represents the averaged RoI feature vectors that have not been selected yet, and $\mathrm{\boldsymbol{q}}_i$ denotes the previously selected RoI feature vectors. The hyperparameter $\lambda$ is introduced to balance the contributions within the scoring function, in which the former term aims to select the most representative one from among those that have not been selected, while the latter term aims to select something different from what has already been selected before. The experiment related to $\lambda$ can be found in Section \ref{subsec:lambda_search}.

CSOD selects data corresponding to the maximum value in Eq. (\ref{eq:submodular_gain}) for each class. If a chosen data point includes multiple classes, the features of these classes are considered part of the previously selected $q_i$. This method systematically cycles through each class, ensuring unique selections, until it reaches the targeted number of choices.
\section{Experiments}
\label{sec:experiments}
In this section, we empirically validate the effectiveness of CSOD through experiments. First, we will show that CSOD outperforms various random selections and other coreset selection methods originally designed for image classification. We will then investigate the tendency associated with the number of selected images and the hyperparameter $\lambda$ of Eq.~(\ref{eq:submodular_gain}). Additionally, we will compare performance when averaging RoI features of a class within an image versus using individual features as they are.
Furthermore, we will extend our analysis to evaluate the performance of different datasets and various network architectures.

\begin{table*}[t]
  \centering
  \begin{tabular}{lccccc} 
  \toprule
    Selection Method & 20 & 100 & 200 & 500 & 1000 \\
  \midrule
    Random (Uniform) &  9.8 $\pm$ 2.2  & 27.9 $\pm$ 1.6 & 37.9 $\pm$ 1.1 & \textcolor{blue}{50.7 $\pm$ 1.0} & \textcolor{blue}{58.4 $\pm$ 0.6} \\
    Herding~\cite{welling2009herding} & 4.1 $\pm$ 0.7 & 17.7 $\pm$ 1.2 & 26.0 $\pm$ 0.8 & 37.8 $\pm$ 0.7 & 46.4 $\pm$ 0.7 \\
    k-Center Greedy~\cite{sener2017active} & 10.0 $\pm$ 1.3 & 21.8 $\pm$ 2.1 & 32.3 $\pm$ 0.9 & 47.4 $\pm$ 1.0 & 55.9 $\pm$ 0.3 \\
    Submodular Function~\cite{iyer2021submodular} & \textcolor{blue}{12.9 $\pm$ 0.9} & \textcolor{blue}{30.5 $\pm$ 1.3} & \textcolor{blue}{38.6 $\pm$ 0.9} & 48.8 $\pm$ 0.7 & 55.8 $\pm$ 0.3 \\
    \textbf{CSOD (Ours)}  & \textcolor{red}{\textbf{14.5 $\pm$ 1.6}}  & \textcolor{red}{\textbf{34.4 $\pm$ 1.0}} & \textcolor{red}{\textbf{44.3 $\pm$ 0.7}} & \textcolor{red}{\textbf{54.1 $\pm$ 0.7}} & \textcolor{red}{\textbf{60.6 $\pm$ 0.4}} \\
  \bottomrule
  \end{tabular}
  \vspace{-1.0mm}
  \caption{Comparison with random and coreset selection for image classification, reporting AP$_{50}$ on the VOC07 test data. We ran all experiments 20 times, with $\pm$ indicating standard deviation. Note that the standard deviation is to show that the performance gap is clear. Herding, Submodular, and CSOD select subsets deterministically, with only network weight initialization affected by random seeds.
}  
  \label{tab:comparision_selection_voc}
\end{table*}

\subsection{Implementation details}
\label{subsec:implementation_details}
We conducted experiments on Pascal VOC 2007+2012 \cite{everingham2007pascal}, using the trainval set for selection and training, and the VOC07 test set for evaluation.
The 
\hj{metric} is Average Precision at IoU 0.5 (AP$_{50}$). For ablation and analysis, we \hj{chose}
200 images from \hj{20 classes},
training for 1000 iterations.
We averaged performance over 20 runs due to the limited number of images.
We \hj{used}
Faster R-CNN-C4 \cite{ren2015fasterRCNN} with ResNet50 \cite{he2016resnet}. For the selection phase, we used the model weight trained on VOC07+12, provided by detectron2 \cite{wu2019detectron2}. 
After selection, a new model was trained on the chosen subset, with a backbone pre-trained on ImageNet~\cite{imagenet}.
For further details, see Section \ref{app_sec:environment_detail} in the supplementary material.



\subsection{Comparision with Other Selections}
\label{sec:comparison_with_others}

\noindent \textbf{Comparison Targets.}\quad Table \ref{tab:comparision_selection_voc} shows the comparison list. Random refers to the method where one image per class is randomly selected in turn. There can be multiple classes in a single image, and when selecting in turn by class, the images were chosen without duplication to make the target number of images.  Additional experiments regarding Random Selection are conducted in Section \ref{subsec:exp_comparision_with_random}.

Coreset selection methods for image classification, such as Herding~\cite{welling2009herding}, k-Center Greedy~\cite{sener2017active}, and Submodular function~\cite{iyer2021submodular}, were also compared. 
The CSOD's \hj{network} weight was employed, and the backbone feature was globally average pooled to utilize that feature vector for selection. Similar to random selection, images were evenly chosen from each class.

\noindent \textbf{Result.}\quad As seen in Table \ref{tab:comparision_selection_voc}, our method consistently shows the highest results. 
Random selection shows higher performance than some existing methods, which implies that these methods are designed only for the image classification task and yield lower results in object detection.
The submodular function showed some effectiveness when the number of data was low. However, as mentioned in Section \ref{sec:intro} \hj{and Figure \ref{fig:overview}}, it is a modeling that fundamentally cannot consider multiple objects of various sizes and locations. Therefore, it not only showed lower results than random when selecting over 500 images but also significantly lower results than CSOD. Based on these results, future experiments for comparison will be conducted with Random selection.

\subsection{Comparison with Random Selections}
\label{subsec:exp_comparision_with_random}
\begin{figure}[t]
    \centering
    \includegraphics[width=0.915\linewidth]{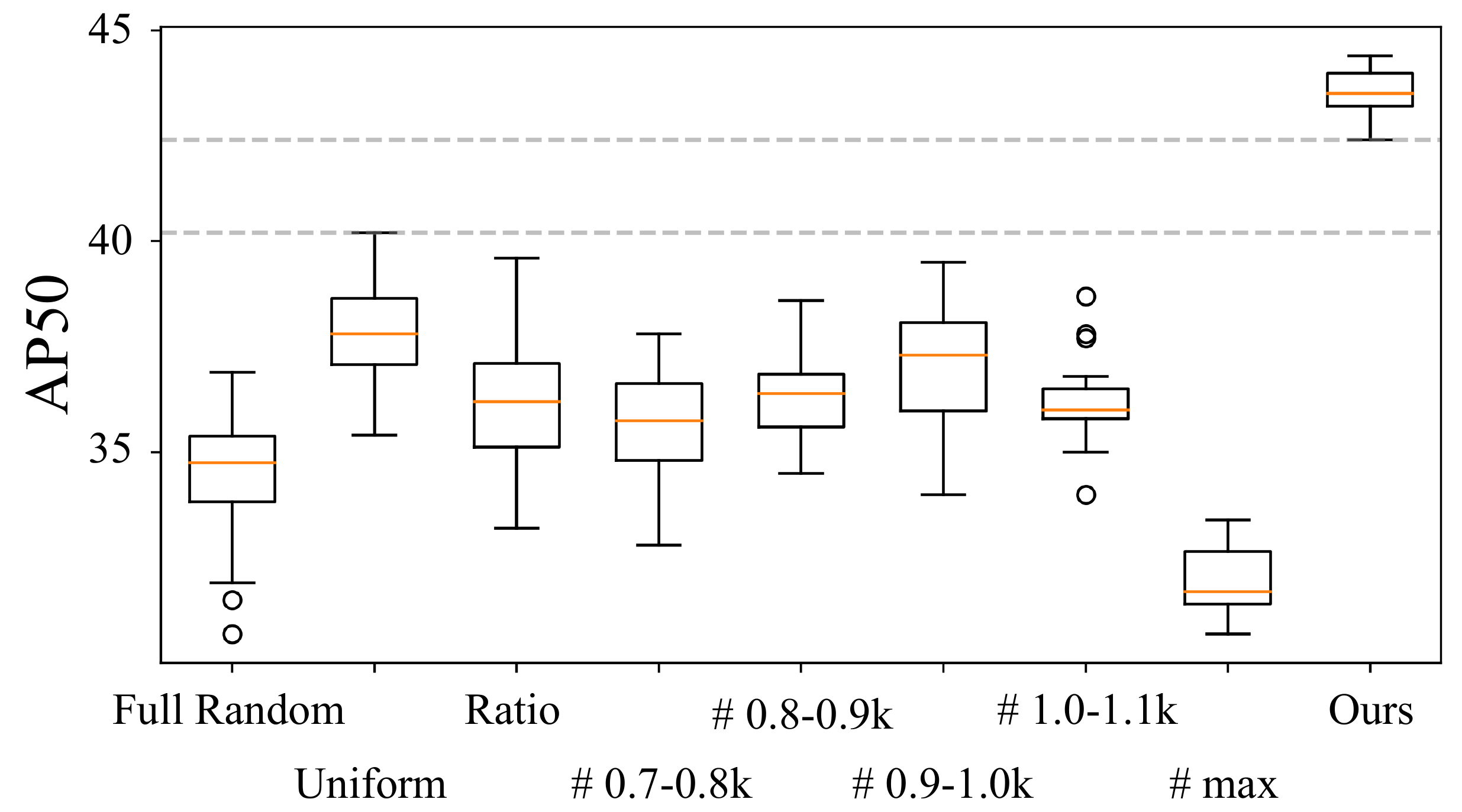}
    \vspace{-0.3mm}
    \caption{Comparison with various selection methods. `$\#$' denotes the number of objects in the selected data.}
    \label{fig:compare_random_ap}
\end{figure}

Figure \ref{fig:compare_random_ap} shows that our approach consistently outperforms other selection methods when selecting 200 images. Notably, in this comparison, ``$\#$ max'' and ``Ours'' are the only methods without randomness, while the rest incorporate some degree of randomness. Therefore, we did not specifically address the performance variance of each selection method. Our method was implemented with a fixed set and without randomness, leading to reduced performance variance only comes from the training process.
This experiment's significance lies in the clearly higher performance of ours compared with those of other methods.

We categorized random selection into several methods. ``Full random'' selects 200 images randomly, but repeats the process if any classes have no objects in those 200 images.
``Uniform'' and ``Ratio'' involve sampling images one by one for each class until 200 images are selected (sampling without replacement). In these cases, images selected from one class are excluded from selection in other classes, as an image can contain objects in multiple classes. ``Uniform'' distributes images evenly with 10 per class, while ``Ratio'' selects images based on the proportion of images per class.

\hj{The} CSOD \hj{result} consists of 1,032 annotations.
Therefore, we also experimented with random selection while controlling for the number of annotations. 
``\# 700-1100'' limits annotations to this range using the Uniform method.
``\# max'' also follows Uniform but selects images based on the annotation count in descending order rather than selecting them randomly.
\begin{figure}[t]
    \centering
    \includegraphics[width=0.915\linewidth]{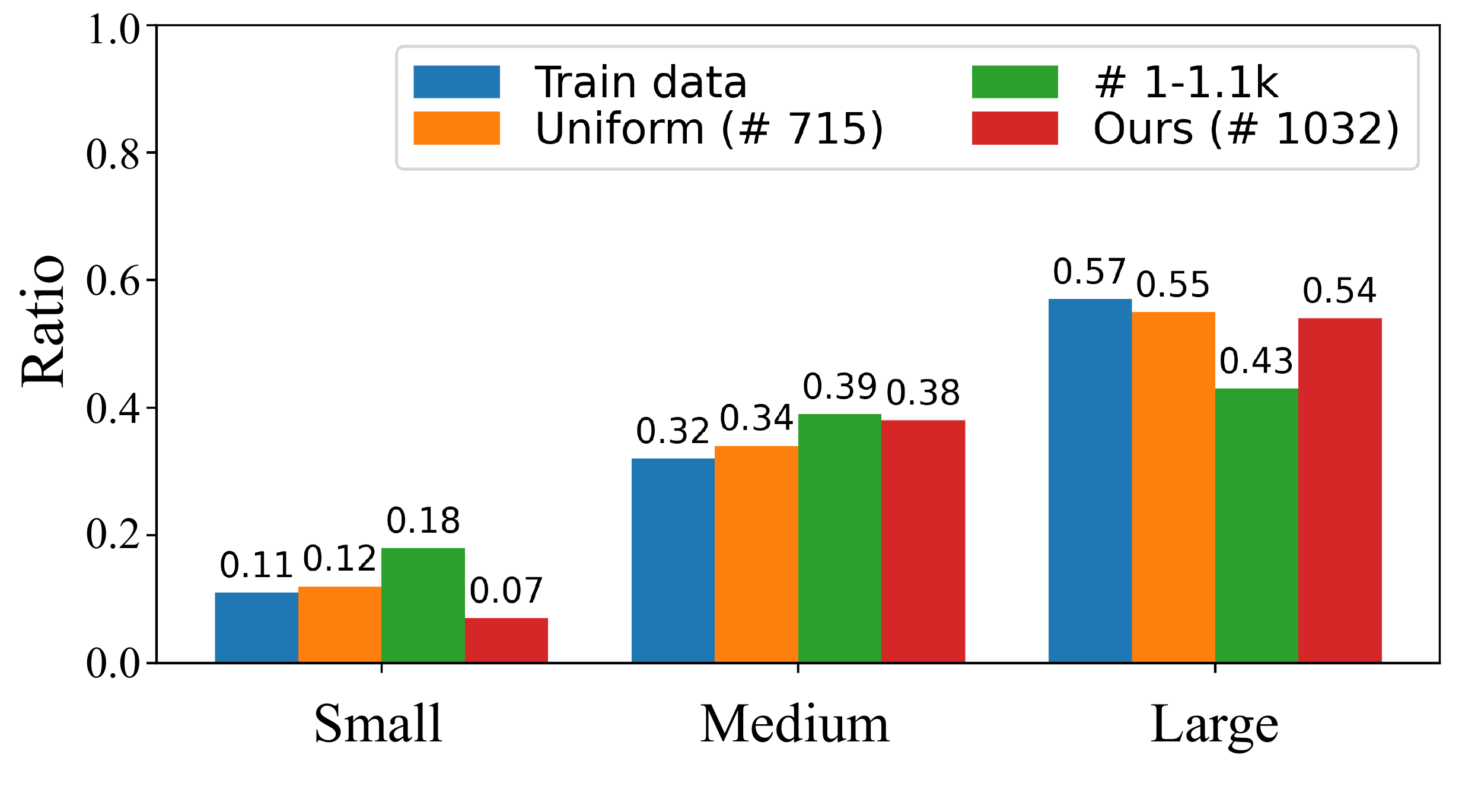}
    \vspace{-0.3mm}
    \caption{Ratio of box sizes. We followed the size criteria provided by VOC. 
    `$\#$' denotes the number of objects in the selected data.}
    \label{fig:compare_random_anno_ratio}
\end{figure}
\subsubsection{The performance and object size ratio.}
\label{subsubsec:object_size_ratio}
Figure \ref{fig:compare_random_anno_ratio} shows the relationship between box size, object count, and performance. 
We conducted two comparisons.
First, we compared Uniform and CSOD.
We observed that the ratio of Uniform was closer to that of the entire dataset in terms of KL divergence, while Ours had more annotations.
Second, we compared CSOD and `\# 1-1.1k'. Both methods had similar object counts, but CSOD's box size ratio was closer to that of the training data.
While we cannot definitively assert causality, it appears that a well-represented subset with an equal number of images has a correlation with both box size and object count.

\subsection{Analysis of the number of images and the hyperparameter \texorpdfstring{$\lambda$}{l}}
\subsubsection{The number of selected images}
\begin{figure}[t]
\centering
\includegraphics[width=0.925\linewidth]{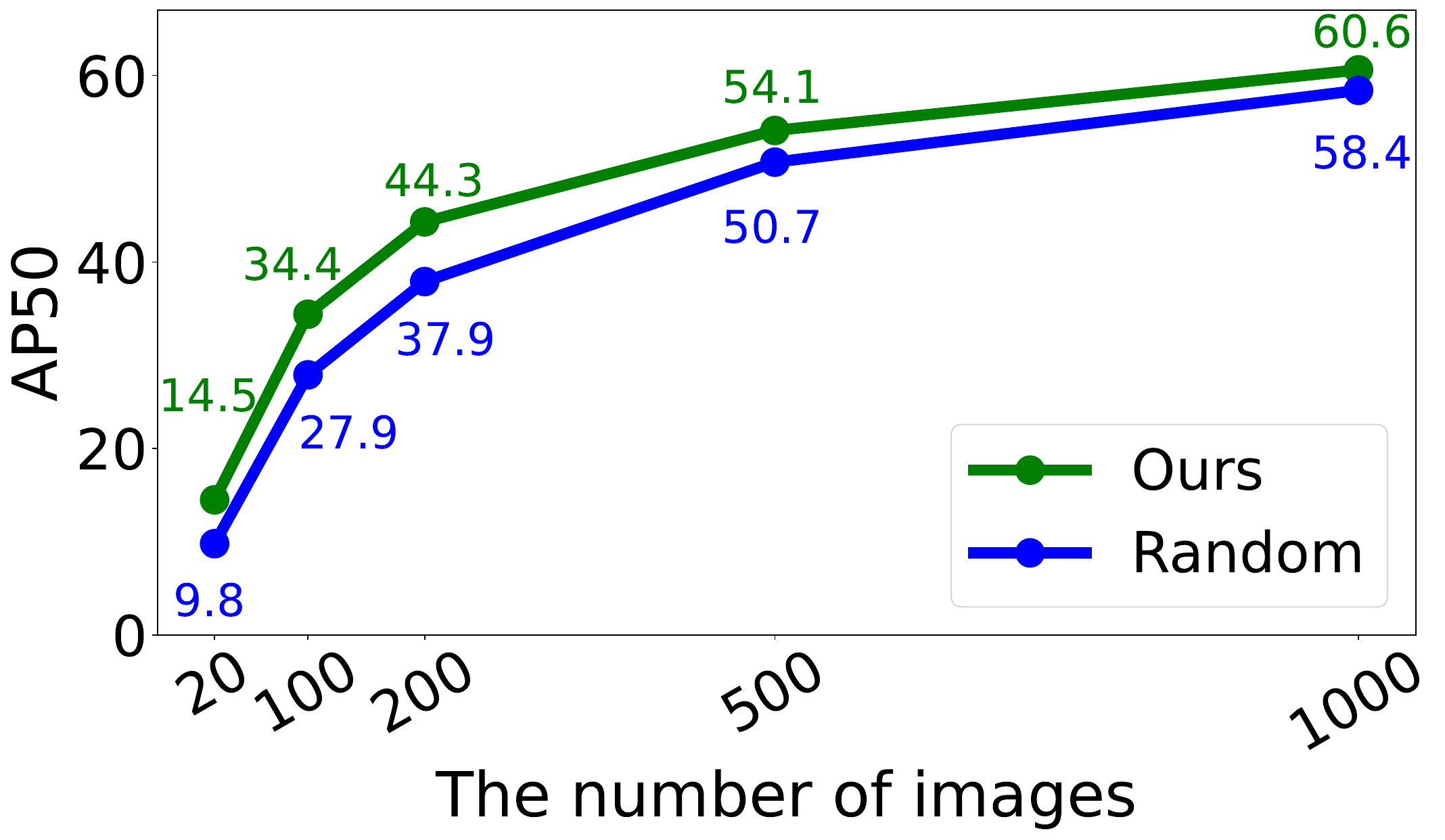}
\vspace{-0.6mm}
\caption{Performance according to the selected image counts.}
\label{fig:num_image}
\end{figure}

Figure \ref{fig:num_image} shows performance with the number of selected images. 
Since selecting 20 images indicates only one image per class is selected, $\lambda$ is meaningless. For other cases (100, 200, 500, and 1000), we set $\lambda$ as (0.0125, 0.04375, 0.0625, and 0.025), respectively.
Compared to the random selection, we observe that as the number of selected images increases, the performance gap naturally decreases, but it consistently remains at a high level.

\begin{figure}[t]
\centering
\includegraphics[width=0.925\linewidth]{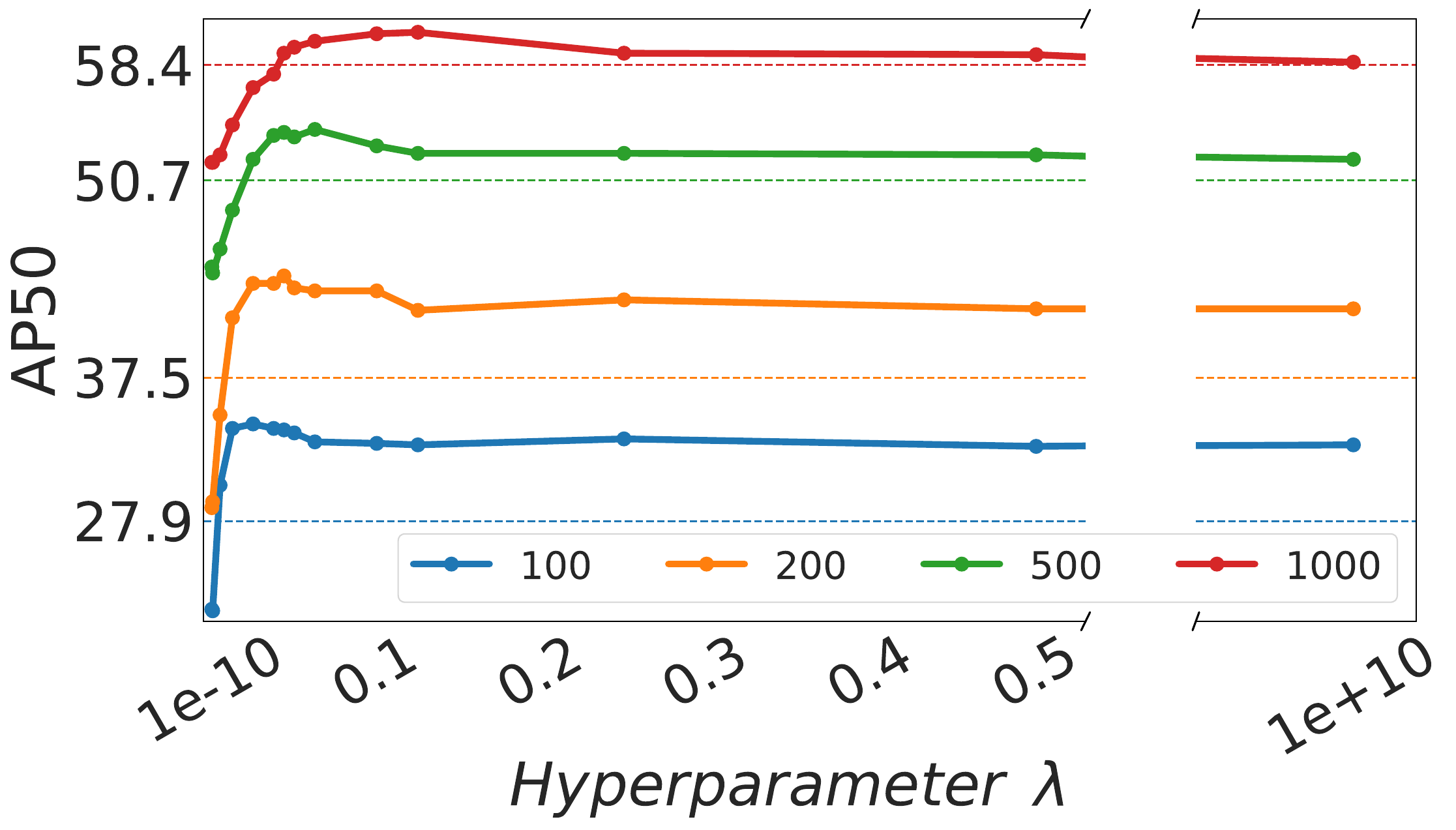}
\vspace{-0.6mm}
\caption{AP$_{50}$ and $\lambda$. Dashed lines represent random selection performance.}
\label{fig:lambda}
\end{figure}

\subsubsection{Balance hyperparameter \texorpdfstring{$\lambda$}{l}}
\label{subsec:lambda_search}
Figure~\ref{fig:lambda} illustrates the relationship between performance and $\lambda$ in Eq.~(\ref{eq:submodular_gain}). 
A high $\lambda$ value (1e+10) means selecting images based solely on cosine similarity, prioritizing representative images.
In contrast, a small $\lambda$ value (1e-10) means selecting an image per class with the highest cosine similarity first and then selecting images that are as dissimilar as possible from those already selected. In other words, it emphasizes diversity from a cosine similarity viewpoint.

The observations can be made: Firstly, our approach outperforms random selection when $\lambda$ is above a certain threshold.
Secondly, it is better to consider both representativeness and diversity by appropriately tuning $\lambda$ rather than simply selecting images purely based on the order of cosine similarity (1e+10). 
Lastly, the optimal $\lambda$ value varies depending on the number of images to be selected, as the greedy selection process (Section \ref{subsec:method_greedy_selection}) progressively increases the number of selected images. Please refer to Table \ref{append_tab:lambda} \hj{in the supplementary material} for the AP$_{50}$ values corresponding to the $\lambda$ values.

\begin{table}[t]
\begin{center}
\resizebox{\columnwidth}{!}{%
\begin{tabular}{cccccccc}
\toprule
&& \multicolumn{5}{c}{The number of changes in the image list} \\ 
\cmidrule{3-7}
&& 0 & 18 & 32 & 45 & 113 \\
\midrule
\multirow{2}{*}{Objectwise} & $\lambda$ & 1e+10 & 0.125 & 0.075 & 0.051 & 0.015 \\
& AP$_{50}$ & 40.4 & 40.3 & 40.7 & 41.4 & 39.0 \\
\midrule
\multirow{2}{*}{Imagewise (Ours)} & $\lambda$ & 1e+10 & 0.100 & 0.050 & 0.038 & 0.013 \\
& AP$_{50}$ & 42.1 & 43.3 & 43.5 & 43.8 & 41.5 \\
\bottomrule
\end{tabular}
}
\end{center}
\vspace{-2.5mm}
\caption{
Comparison with two methods, ``Imagewise'' (averaging RoI vectors) and ``Objectwise'' (not averaging), for selecting 200 images.
The number of changes in the image list is based on $\lambda=1e+10$ as the reference point (The smaller $\lambda$, the severer the change). $\lambda$ is rounded to the fourth decimal place.
}
\label{tab:comparison_with_objwise} 
\end{table}

\subsection{Effectiveness of Averaging RoI feature vectors}
\subsubsection{Performance comparison}
Table \ref{tab:comparison_with_objwise} compares performance between averaging the RoI feature vectors of the same class (Imagewise) or not (Objectwise). These two cases have different balance strengths for $\lambda$, as Imagewise averages within the same class, resulting in significantly fewer RoI feature vectors. Therefore, we compared the extent to which the image list changes, using 1e+10 as the reference point. Remarkably, we observed consistent high performance regardless of $\lambda$ values. However, even Objectwise outperformed random selection, achieving a result higher than 37.5 of random selection.

\begin{table}[t]
\begin{center}
\resizebox{0.93\columnwidth}{!}{%
\begin{tabular}{ccccccc}
\toprule
& 20 & 40 & 60 & 80 & 100 & 200 \\ 
\midrule
Objectwise & 13.0 & 20.6 & 25.8 & 29.1 & 31.5 & 40.4 \\ 
Imagewise (Ours) & 14.2 & 23.1 & 27.3 & 30.1 & 32.9 & 42.1 \\ 
\bottomrule
\end{tabular}
}
\end{center}
\vspace{-2.3mm}
\caption{$\lambda$ is 1e+10 in all cases, meaning that we selected based solely on cosine similarity ranking.}
\label{tab:num_img_and_objwise}
\end{table}
\subsubsection{Representativeness: Objectwise vs. Imagewise feature vector}
Table \ref{tab:num_img_and_objwise} compares Objectwise and Imagewise based on the number of selected images when $\lambda$=1e+10. This experiment highlights that even if the cosine similarity of a single object within an image is exceptionally high, that image may not effectively represent the overall distribution of the data. 
For example, the case where only one image per class is selected (20 in the table) indicates how well a single image represents the corresponding class.
The table shows the superiority of our imagewise selection over objectwise selection.

\subsubsection{Visulization of objectwise selection}
\begin{figure}[t]
    \centering
    \includegraphics[width=0.90\linewidth]{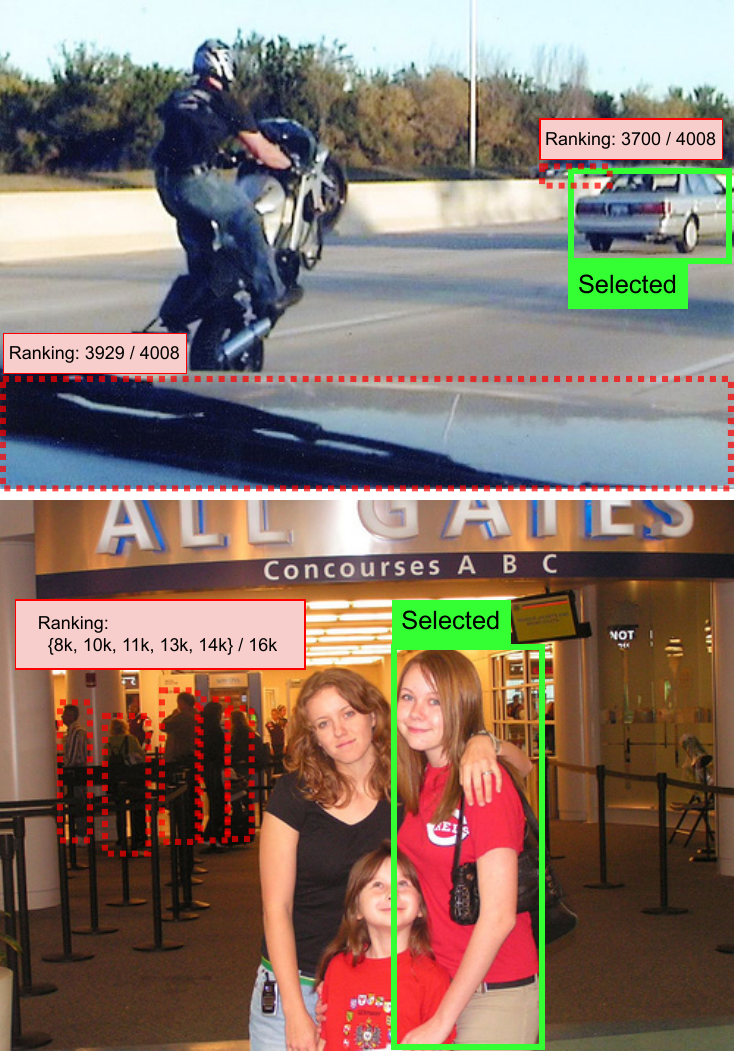}
    \vspace{-0.5mm}
    \caption{Examples of the Objectwise selection. Top: `car' class.  Bottom: `person' class. The red dotted boxes indicate objects with low rankings in Eq. (\ref{eq:submodular_gain}).}
    \label{fig:vis_objwise_weakness}
\end{figure}

Figure \ref{fig:vis_objwise_weakness} illustrates a limitation of the objectwise approach, where the selected image may not effectively represent the entire dataset.
Even if an image is selected because it contains an object with high cosine similarity, it does not guarantee that other objects within the same image will have similarly high cosine similarities. In other words, the cosine similarity of one object in an image with all the other objects in the entire dataset may not accurately represent the cosine similarities of all objects in that image.

\subsubsection{Why imagewise (averaging) selection over 
objectwise selection?}

Let us compare object counts and size ratios in Section \ref{subsubsec:object_size_ratio}. In our Imagewise approach, there are 1,032 objects in our 200 selected images, which is higher than 806 in the Objectwise approach. Additionally, when considering the size ratios (small, medium, large), the Imagewise approach results in (7.3\%, 38.2\%, 54.5\%), which is closer to the large object ratio in the entire train data, (10.5\%, 32.0\%, 57.5\%), compared to the Objectwise, (10.2\%, 37.2\%, 52.6\%). When we calculated the KL divergence between the distributions of the selected images and the entire training data, we found that Objectwise had a KL divergence of 0.006, lower than Imagewise's 0.013. 

This suggests that the number of annotations played a more significant role in the performance than the size ratio in the case of Imagewise and Objectwise. Despite the higher KL divergence for Ours, there were substantial differences in the number of annotations for each size. Ours had counts of (75, 395, 562) for each size, whereas Objectwise had counts of (82, 300, 424).

We formulated a hypothesis that ``As the number of objects within an image increases and their sizes are larger, the cosine similarity between the class's entire average RoI vector (class prototype) and the image's average RoI vector (image prototype) for that class will be higher.''

\begin{table}[t]
\begin{center}
\resizebox{0.75\columnwidth}{!}{%
\begin{tabular}{c|ccc}
\toprule
Object count & 1 & 2-4 & 5 or more \\ 
\midrule
Large & 0.866 & 0.931 & 0.939  \\ 
Medium & 0.862 & 0.915 & 0.931  \\ 
Small & 0.798 & 0.818 & 0.831  \\ 
\bottomrule
\end{tabular}
}
\end{center}
\vspace{-2.3mm}
\caption{Cosine similarity between the entire average features and the average feature of each image by size in \hj{the} `person' class.}
\label{tab:distance_from_centroid}
\end{table}

To validate this hypothesis, we conducted the experiment presented in Table \ref{tab:distance_from_centroid}. Initially, we averaged all RoI vectors for the `person' class (class prototype). Then, we made an averaged RoI vector by size within each image (imagewise-sizewise prototype). We subsequently computed the cosine similarity between the class prototype and the imagewise-sizewise prototypes. The results confirmed that the Imagewise approach leads to a higher selection of larger objects, resembling the entire dataset.

\subsection{Evaluation on the BDD100k dataset}
\label{sec:bdd_result}
BDD100k \cite{yu2020bdd100k} is a significant dataset for autonomous driving, consisting of 100k images, 1.8M annotations, and 10 different classes. Following the official practice, we split the
dataset into 70k for training with 1.3M annotations. Table \ref{tab:bdd} shows the results on the validation data, illustrating that our method consistently achieves higher AP$_{50}$, AP$_{75}$, and AP compared to random selection. Notably, similar performance improvements were observed in our experiments with the VOC dataset. However, BDD100k, explicitly designed for real-world autonomous driving applications, offers a more challenging and realistic benchmark. The fact that our method excels even in the challenging environment of BDD100k further demonstrates its effectiveness and practicality. For implementation details, kindly refer to Section \ref{app_sec:bdd_exp_detail} \hj{in the supplementary material}.

\begin{table}[t]
\centering
\resizebox{0.69\columnwidth}{!}{%
\begin{tabular}{ccccc}
\toprule
 & num img & AP$_{50}$ & AP$_{75}$  & AP \\
\midrule
\multirow{3}{*}{200} & Random  & 25.8 & 9.7 & 12.0\\
 & Ours  & 29.0 & 10.8 & 13.5\\ 
 & $\Delta$  & +3.2 & +1.1 & +1.5\\
\midrule
\multirow{3}{*}{500} & Random  & 32.2 & 13.4 & 15.8\\
& Ours & 35.1 & 14.9 & 17.5 \\
& $\Delta$  & +2.9 & +1.5 & +1.7\\
\midrule
\multirow{3}{*}{1000} & Random  & 37.1 & 16.2 & 18.6\\
& Ours  & 39.4 & 17.8 & 20.1\\ 
& $\Delta$ & +2.3 & +1.6 & +1.5\\
\midrule
\multirow{3}{*}{2000} & Random  & 42.1 & 19.7 & 22.0\\
& Ours  & 43.7 & 21.0 & 23.2\\
& $\Delta$ & +1.6 & +1.3 & +1.2 \\
\bottomrule
\end{tabular}
}
\vspace{-1.0mm}
\caption{BDD100k result}
\label{tab:bdd}
\end{table}

\subsection{Evaluation on the COCO dataset}
The COCO2017 dataset \cite{lin2014microsoft} encompasses 80 classes, partitioned into 118K images for the training set and 5K for the validation set. 
Experiments were conducted with subsets of 400 and 800 images selected from the training data. Table \ref{tab:coco} shows the outcomes for the validation set. Despite the substantially larger size of the dataset compared to VOC, it was observed that our method was effective. For reproducibility please refer to Section \ref{app_sec:coco_exp_detail} \hj{in the supplementary material}.

The BDD dataset has more annotations despite having fewer images compared to COCO. Interestingly, the performance improvement margin on COCO was smaller than on BDD. This observation raises several points for consideration. The BDD dataset, focused on autonomous driving, predominantly encompasses outdoor scenes with less diversity, such as perspective, compared to COCO. Conversely, COCO spans a wider spectrum, including both indoor and outdoor scenes and a larger variety of classes. This diversity potentially renders the accurate representation of the entire data distribution with a subset of images more challenging. This observation not only clarifies our current results but also highlights this as a key area for future study.

\begin{table}[t]
\centering
\resizebox{0.69\columnwidth}{!}{%
\begin{tabular}{ccccc}
\toprule
 & num img & AP$_{50}$ & AP$_{75}$ & AP \\
\midrule
\multirow{3}{*}{400} & Random & 15.1 & 4.9  & 6.6\\
 & Ours & 16.7 & 5.8  & 7.5\\
 & $\Delta$ & +1.6 & +0.9 & +0.9\\
\midrule
\multirow{3}{*}{800} & Random & 19.4 & 7.5  & 9.1\\
& Ours & 20.1 & 8.2  & 9.6\\
& $\Delta$ & +0.7 & +0.7 & +0.5\\
\bottomrule
\end{tabular}
}
\vspace{-1.0mm}
\caption{COCO2017 result}
\label{tab:coco}
\end{table}

\subsection{Cross-architecture evaluation}
\label{subsec:exp_cross_arch}
Table \ref{tab:cross_archi} presents an experiment in which we assessed whether the 500 images selected using Faster R-CNN remained effective for different networks, namely RetinaNet \cite{lin2017focal} and FCOS \cite{tian2019fcos}. We were able to confirm the effectiveness of images selected with Faster R-CNN for other networks as well. Unlike Faster R-CNN, these two networks often encountered training issues due to loss explosion when following their respective default hyperparameters.
Therefore, we adjusted the hyperparameters, such as the learning rate and gradient clipping, but it is important to note that the hyperparameters for random selection and our method remained consistent. Please refer to Section \ref{app_sec:cross_archi_detail} \hj{in the supplementary material} for reproducibility. 

\begin{table}[t]
\centering
\begin{tabular}{ccc}
\toprule
 & RetinaNet & FCOS \\
\midrule
random & 54.5 & 47.9 \\
ours & 58.3 & 53.1 \\
$\Delta$ & +3.8 & +5.2 \\
\bottomrule
\end{tabular}
\vspace{-0.5mm}
\caption{Cross architecture experiment. We trained the models on 500 VOC images and reported AP$_{50}$.}
\label{tab:cross_archi}
\vspace{-0.5mm}
\end{table}
\section{Discussion}
\noindent\textbf{Conclusion.} \quad 
We have proposed a Coreset selection method for Object Detection tasks\hj{, addressing the unique challenges presented by multi-object and multi-label scenarios. This stands in contrast to traditional image classification approaches.}
Our approach considers both representativeness and diversity while taking into account the difficulties we have outlined \hj{in Section \ref{sec:intro} and illustrated in Figure \ref{fig:overview}}. Through experiments, we have demonstrated the effectiveness of our method, and its applicability to various architectures. We hope this research will further develop and find applications in diverse areas, such as dataset distillation.

\noindent\textbf{Limitation.} \quad
While our research leveraged RoI features from ground truth boxes and achieved promising results, it is important to note certain limitations. Firstly, we did not explicitly incorporate background features, which could provide additional context and potentially enhance coreset selection in object detection. Future research could explore the explicit utilization of background features. Our approach, which selects greedily on a class-by-class basis, can take into account the RoI features of the current class even when they were selected during the turn of other classes. However, our method does not simultaneously incorporate the features of other classes within the same image. Further research could explore ways to capture interactions between different classes more effectively within a single image. 

\noindent\textbf{Future work.} \quad Since CSOD considers localization, there may be aspects that can be applied to other tasks related to localization, such as 3D object detection. Furthermore, while dataset distillation has predominantly been studied in the context of image classification, it could also become a subject of research in the field of object detection datasets.
\\
\\
\noindent\textbf{Acknowledgements.} This work was supported by NRF grant (2021R1A2C3006659) and IITP grants (2022-0-00953, 2021-0-01343), all funded by MSIT of the Korean Government.

{
    \small
    \bibliographystyle{ieeenat_fullname}
    \bibliography{main}
}

\newcommand{\supplementstart}{
    \counterwithin*{section}{part} 
    \counterwithout{figure}{section} 
    \counterwithout{table}{section}  
    \setcounter{section}{0}
    \renewcommand{\thesection}{S\arabic{section}}
    \setcounter{figure}{0}
    \renewcommand{\thefigure}{S\arabic{figure}}
    \setcounter{table}{0}
    \renewcommand{\thetable}{S\arabic{table}}
}

\supplementstart

\clearpage
\setcounter{page}{1}
\maketitlesupplementary

\begin{figure*}[t]
    \centering
    \includegraphics[width=0.8\linewidth]{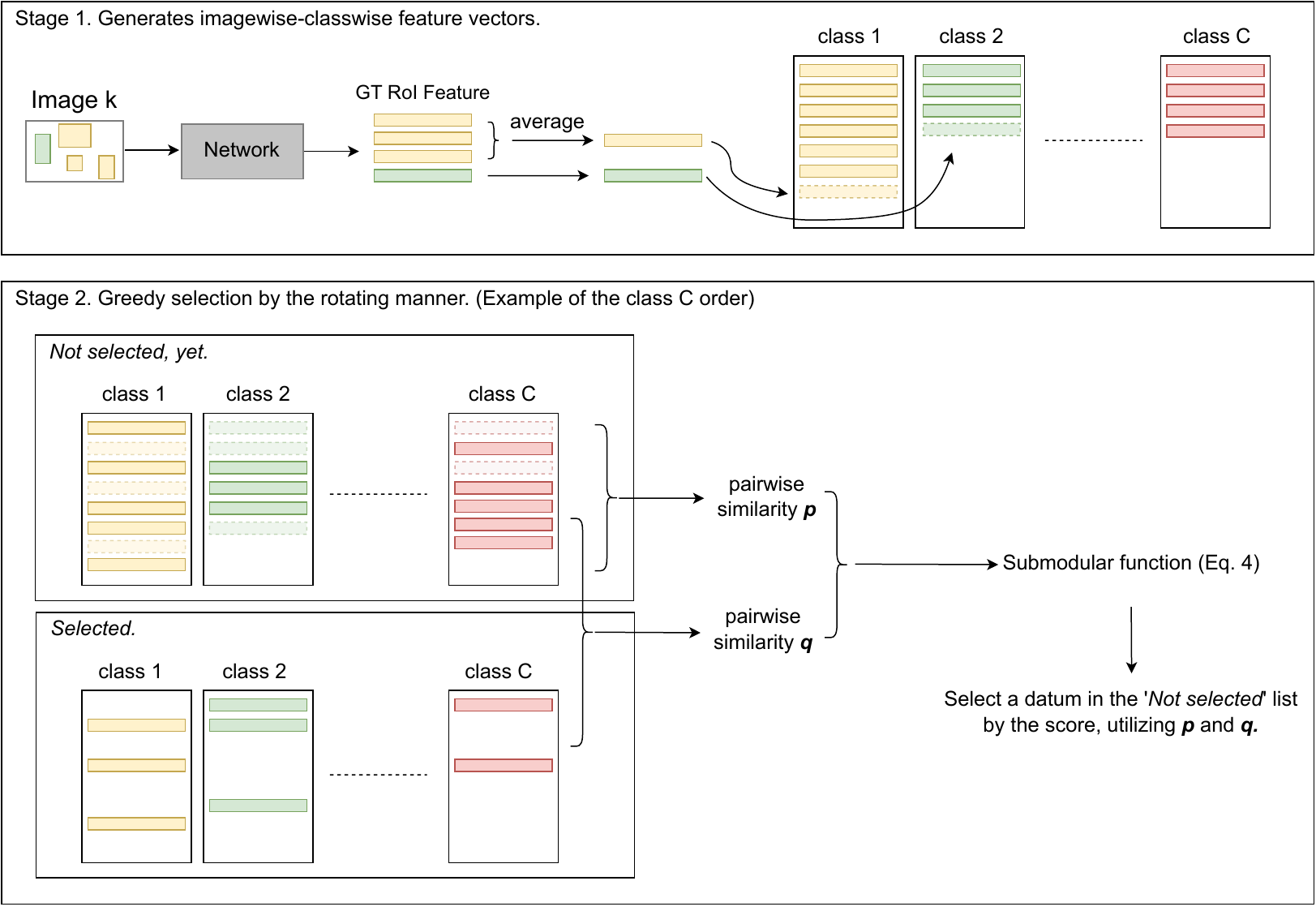}
    \caption{Stage 1 is to create imagewise-classwise feature vectors by applying RoI pooling to the ground truth boxes and subsequently averaging them by class. In Stage 2, illustrated by class C as an example, data selection for each class is performed. It calculates scores that consider the balance between previously selected and the not-selected data (Eq. \ref{eq:submodular_gain}). After selection by Eq. \ref{eq:submodular_gain}, the datum selected in the current selection step is popped from the `\textit{Not selected, yet}' list and inserted into the `\textit{Selected}' list. After selecting a datum for class C, the process returns to class 1 and repeats the same steps until the desired number of images is reached.}
    \label{append_fig:overview}
\end{figure*}

\begin{table*}[t]
\begin{center}
\resizebox{0.98\textwidth}{!}{%
\begin{tabular}{c|ccccccccccccccc}
\toprule
count & 1e-10 & 0.0005 & 0.005 & 0.0125 & 0.025 & 0.0375 & 0.04375 & 0.05 & 0.0625 & 0.1 & 0.125 & 0.25 & 0.5 & 1e+10 \\
\midrule
100 & 22.0 & 21.9 & 30.3 & 34.1 & \textcolor{red}{34.4} & 34.1 & 34.0 & 33.8 & 33.2 & 33.1 & 33.0 & 33.4 & 32.9 & 33.0 \\
200 & 28.8 & 29.2 & 35.0 & 41.5 & 43.8 & 43.8 & \textcolor{red}{44.3} & 43.5 & 43.3 & 43.3 & 42.0 & 42.7 & 42.1 & 42.1 \\
500 & 44.9 & 44.5 & 46.1 & 48.7 & 52.1 & 53.7 & 53.9 & 53.6 & \textcolor{red}{54.1} & 53.0 & 52.5 & 52.5 & 52.4 & 52.1 \\
1000& 51.9 & 51.9 & 52.4 & 54.4 & 56.9 & 57.8 & 59.2 & 59.6 & 60.0 & 60.5 & \textcolor{red}{60.6} & 59.2 & 59.1 & 58.6 \\
\bottomrule
\end{tabular}
}
\end{center}
\vspace{-2.0mm}
\caption{AP$_{50}$ on Pascal VOC. The optimal $\lambda$ increases as the number of images grows.}
\label{append_tab:lambda}
\end{table*}

\section{Additional implementation details}
\label{app_sec:environment_detail}
Section \ref{subsec:implementation_details} describes implementation details, and here we introduce additional implementation details.
During training with the selected data, we used the SGD optimizer with a learning rate of 0.02, weight decay of 0.0001, and momentum of 0.9. The number of iterations was as follows:
When the number of images was 200 or fewer, we trained the initialized network for 1000 iterations.
For 500 images, we trained it for 2000 iterations.
With 1000 images, we trained it for 4000 iterations.
When there were 200 images or fewer, we performed training on the initialized network for 1000 iterations to ensure loss convergence. The learning rate was reduced to 0.1 times the initial value at 80\% of the training iterations. The warm-up was conducted for 100 iterations, and there was no gradient clipping.

Regarding image sizes, we resized the shorter side to 800 pixels and maintained the original aspect ratio, ensuring the longer side remained below 1333 pixels. During training, resizing was done within the range of (480-800) with a step size of 32.

\section{BDD100k experiment detail}
\label{app_sec:bdd_exp_detail}
Like Pascal VOC, we experimented with Faster R-CNN-C4 with ResNet50. For the VOC experiment, we used the Faster R-CNN weight provided by detectron2. However, for the BDD experiment, there was no weight provided by detectron2. Therefore, we used ImageNet pretrained backbone to train the Faster R-CNN on the train data. We trained with a max iteration of 90k and a learning rate decay at 60k and 80k. The rest, like SGD, learning rate, and data augmentation strategy, was the same with VOC.

We initially sampled 5,000 images for each class and then ran our method. For more on this setting, please refer to Section \ref{app_sec:selection_after_sampling}. 
For the balance hyperparameter $\lambda$ of Eq. (\ref{eq:submodular_gain}), we set them to (0.075, 0.0675, 0.1, 0.1875) for the images (200, 500, 1000, 2000), respectively. 
After selection, we trained the selected subset during 4k iteration for 200 images, 4k iteration for 500 images, 8k iteration for 1000 images, and 8k iteration for 2000 images. And, the learning rate was decayed at 80\% of the max iteration.

\section{COCO2017 experiment detail}
\label{app_sec:coco_exp_detail}
Like Pascal VOC, we experimented with Faster R-CNN-C4 with ResNet50. Similar to VOC, we used the Faster R-CNN weight provided by detectron2 for selection.

We initially sampled 30,000 images for each class and then ran our method. For more on this setting, please refer to Section \ref{app_sec:selection_after_sampling}. 
For the balance hyperparameter $\lambda$ of Eq. (\ref{eq:submodular_gain}), we set them to (400, 800) for the images (0.05, 0.1), respectively. 
After selection, we trained the selected subset during 4k iteration for 400 images and 8k iteration for 800 images. And, the learning rate was decayed at 80\% of the max iteration.

\section{Cross architecture detail}
\label{app_sec:cross_archi_detail}
In the case of RetinaNet and FCOS, as mentioned in Section \ref{subsec:exp_cross_arch}, following the configurations of each detector or the configurations we experimented with in Faster R-CNN resulted in significant training instability due to loss explosion. Therefore, when training with 500 images, we followed these hyperparameters:
We increased the number of training iterations from 2000 to 6000, matching the learning rate decay point at 5,200 iterations accordingly.
The learning rate for RetinaNet was set to 0.01, following Detectron2, while for FCOS, it was reduced from 0.01 to 0.005.
We extended the warm-up iteration from 100 to 1000 iterations.
Gradient clipping was introduced with a threshold of 1.0, which was previously absent. 
We reduced the image size in training and testing, scaling down the shorter side from 800 to 600 pixels and the longer side from 1333 to 1000 pixels.
We also reduced the data resize augmentation range from 480-800 to 360-600.
For other network hyperparameters, we followed Detectron2's settings for RetinaNet and used the official code's configurations for FCOS.

\section{Sampling first, then selection for reducing cosine similarity calculation}
\label{app_sec:selection_after_sampling}

Table \ref{tab:sampling} presents the experiment that we first reduced the number of data per class by random selection and then applied our method. Note that the numbers per class are not precisely equal, as some images might contain multiple classes, but rather approximations. We conducted this experiment for two reasons: handling classes with a large number of images, which can make cosine similarity calculations time-consuming, and addressing class imbalance. 

Our results show a clear trend. Even with just 50 images per class, our algorithm outperformed random selection by a significant margin.
Moreover, increasing the sample size to 1,000 images per class yielded similar performance.  This implies that, even when dealing with classes with an unusually high number of images, as long as computational resources allow, sampling without replacement is expected to provide markedly better results than random selection.
For reference, the minimum, maximum, and average number of images per class are 423, 6,469, and 1,281, respectively.

\begin{table}[t]
\begin{center}
\resizebox{\columnwidth}{!}{%
\begin{tabular}{cc|ccccccc|c}
\toprule
 & \multirow{2}{*}{Random} & \multicolumn{7}{c|}{The number of sampled image per class} & \multirow{2}{*}{Ours} \\
 & & 50 & 100 & 200 & 300 & 400 & 500 & 1000 &  \\
\midrule
AP$_{50}$ & 37.5 & 42.0 & 42.8 & 43.6 & 43.9 & 43.9 & 43.5 & 44.0 & 44.3 \\
\bottomrule
\end{tabular}
}
\end{center}
\vspace{-2.0mm}
\caption{Coreset Selection after Sampling. In all cases, we ultimately selected 200 images.}
\label{tab:sampling}
\end{table}

\section{Analysis of the ratio of classwise annotation counts}
\begin{figure*}[t]
    \centering
    \includegraphics[width=0.95\linewidth]{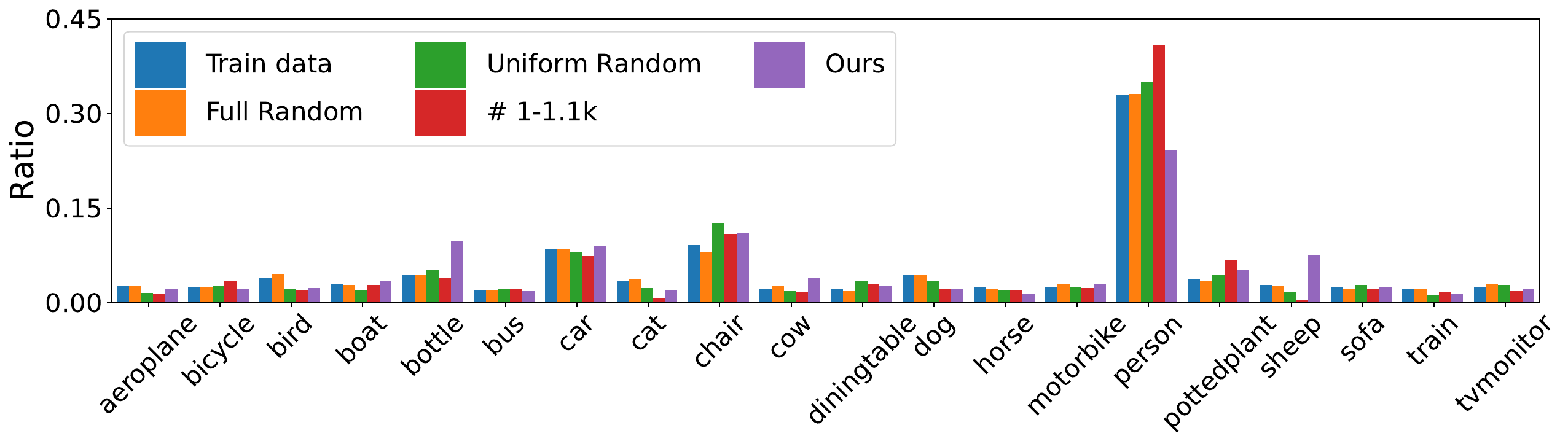}
    \caption{Annotation ratios for each class. When considering the KL divergence as the criterion with train data, Full Random was the closest. However, when calculating the balance level by computing the entropy, Ours showed the most balance.}
    \label{append_fig:classwise_annotation}
\end{figure*}

We explored how selection methods affect the class balance of annotations. Figure \ref{append_fig:classwise_annotation} shows the results. Note that in this analysis, we considered the proportions without considering the number of annotations or object sizes.

When we calculated the KL divergence with the train data, Full Random appeared to be the closest. We also evaluated how balanced the class ratios were based on the entropy. Our method showed the best balance in terms of the entropy.

Intuitively, we might expect the performance to be better when the subset's class ratios are closest to those in the train data. However, this was not necessarily the case, suggesting two possibilities. First, the number of annotations in the subsets is significantly smaller than in the train data. Therefore, having a relatively balanced dataset, even if it differs from the train data's class ratios, could be more beneficial for learning. 
Second, as shown in Section \ref{subsubsec:object_size_ratio}, the size of objects or the number of annotations might significantly impact performance.


\section{Coreset selection with gradients instead of RoI features}

In our pursuit of selecting an informative subset for training, we conducted an experimental comparison where we explored the use of gradient vectors derived from RoI feature vectors instead of directly employing the RoI feature vectors. This approach was motivated by \cite{guo2022deepcore}, where they demonstrated the effectiveness of gradients in image classification tasks.

The gradient vectors were obtained by backpropagating from the classification loss. Furthermore, we performed our method with the gradient of RoI feature vectors instead of RoI feature vectors. We focused solely on the gradients of classification loss because they have significantly higher dimensions than RoI feature vectors, mainly due to the multiplication of the number of classes. 

In our experiments, when selecting 200 images with $\lambda$ set to 0.05, we compared the AP$_{50}$ values. Results revealed that utilizing our method with RoI feature vectors achieved an AP$_{50}$ of 43.5, whereas when using gradients, the AP$_{50}$ was 42.3.

\section{Class-specific hyperparameter \texorpdfstring{$\lambda$}{l} set differently for each class}
We experimented with setting $\lambda$ of Eq. (\ref{eq:submodular_gain}) differently for each class. Because the number of objects varies among classes, it leads to varying scales in the former term of the right-hand side of Eq. (\ref{eq:submodular_gain}).

So, if we denote the number of objects in class $c$ as $N_c$, we conducted hyperparameter tuning for $\lambda_c$, such as $\lambda_{c} \propto 1/N_c$, $\lambda_{c} \propto 1/\log_{2}(N_c)$, or $\lambda_{c} \propto 1/a^{N_c}$, where $a$ is a hyperparameter. However, we observed no significant difference in AP$_{50}$. Furthermore, class-specific AP$_{50}$ values did not reveal any consistent trends concerning the number of objects.

There are several possible explanations to consider. First, our method selects classes independently, without considering other classes. However, as mentioned in Section \ref{sec:intro}, a single image can contain multiple classes. Therefore, the assumption of complete independence among classes may not hold, and $\lambda_c$ may have indirectly influenced other classes. Second, the VOC dataset exhibits a relatively less severe class imbalance compared to other datasets \cite{oksuz2020imbalance}. 
Based on our observations that there is not a significant performance difference within a specific range of $\lambda$ values, it is possible that the imbalance is not severe.

Data imbalance is a challenging yet crucial issue addressed in many domains. Coreset selection from imbalanced data is also an area that deserves deeper exploration in the future.

\section{What would happen if we discarded all the selected images and chose again?}

We observed an AP${_{50}}$ of 43.5 when we set $\lambda$ to 0.05 and selected 200 images, with 10 images per class in the VOC dataset. Subsequently, we conducted an additional experiment in which we excluded the originally chosen 200 images and then selected another 200 images. In this case, the AP${50}$ value was 42.2.

Two crucial observations emerged from these results: Firstly, the initial selection of 200 images effectively represented the entire dataset. Secondly, even when we reselected 200 images from the remaining dataset after discarding the initial selection, the performance remained significantly superior to random selection, which yielded an AP$_{50}$ of 37.5.

\section{Measurement of CSOD's selection Time}
Table \ref{tab:selection_time} presents the results of measuring the time it takes for CSOD to select data. The time taken to select VOC in Section \ref{sec:comparison_with_others} was measured, and the time for selecting BDD in Section \ref{sec:bdd_result} was measured. 
As the results indicate, there is a tendency for the time to increase linearly with the number of selected images. And, comparing BDD and VOC, the more data there is, the longer it takes to make a selection. This linear increase in time is not a drawback but rather an expected and manageable aspect of the process, ensuring a thorough and proportional selection as the volume of data scales up.

\begin{table}[t]
\begin{center}
\begin{tabular}{c|cccc}
\toprule
 \multirow{2}{*}{Dataset} & \multicolumn{4}{c}{The number of selected images} \\
 & \hspace*{1.5mm} 200 & \hspace*{1.5mm} 500 & \hspace*{1.5mm} 1000 & \hspace*{1.5mm} 2000 \\
 \midrule
VOC & \hspace*{1.5mm} 15 & \hspace*{1.5mm} 42 & \hspace*{1.5mm} 80 & \hspace*{1.5mm} 158 \\
BDD & \hspace*{1.5mm} 310 & \hspace*{1.5mm} 712 & \hspace*{1.5mm} 1393 & \hspace*{1.5mm} 2824 \\
\bottomrule
\end{tabular}
\end{center}
\caption{Selection time measurement result.
When measuring, it was measured using CPU, not GPU. The unit is second.}
\label{tab:selection_time}
\end{table}

\section{Comparative and adaptation study utilizing active learning for object detection}
Talisman \citep{kothawade2022talisman} applied submodular functions in the context of active learning. Talisman focuses on maximizing the performance of rare classes in active learning for object detection. Therefore, when provided with both labeled and unlabeled data, their goal is to select unlabeled data that maximizes the information about rare objects within the labeled data. While our objectives and methods are entirely different from the paper, we share a commonality in the use of submodular functions.

The main differences in the procedures of Talisman and CSOD are as follows:
First, Talisman starts by randomly selecting $N$ objects from each category, while CSOD does not follow this initial step.
Second, Talisman searches for images containing at least one object that maximizes the information given the initial set of $N$ objects. This process uses a method known as a submodular function. It's important to understand that Talisman selects these images based on the presence of a single object that greatly enriches the information content, regardless of how similar or dissimilar the other objects in the image are. 

Kindly note that this succinct overview provides only a glimpse of the procedural differences, and it's crucial to recognize that the two works are guided by distinct objectives, leading to fundamentally different frameworks. Nevertheless, we adapted Talisman to our problem formulation and conducted an experiment to select 200 images from Pascal VOC.

We made several modifications to tailor the Talisman method to our needs: 
\begin{itemize}
\item While Talisman primarily focused on rare classes, we extended its application to all classes in our problem. 
\item Instead of generating RoI features from unlabeled data using RPN outputs, we changed it for creating RoI features from Ground Truth (GT) boxes, given our supervised setting. 
\item The Talisman algorithm initially started with a few labeled rare objects. Our adaptation altered the method to start with three random images per class from the entire dataset. For a fair comparison, CSOD also started with the first three images identical to Talisman.
\end{itemize}
 
As a result, AP$_{50}$ of our adaptation of Talisman was 39.3. This result was higher than the random selection of 37.5 but lower than 43.1 of ours, which started with the first three images identical to Talisman.

\end{document}